\documentclass[letterpaper]{article} 
\usepackage{aaai2026}  
\usepackage{times}  
\usepackage{helvet}  
\usepackage{courier}  
\usepackage[hyphens]{url}  
\usepackage{graphicx} 
\usepackage{natbib}  
\usepackage{caption} 
\usepackage{algorithm}
\usepackage{algorithmic}
\usepackage{amsmath}
\usepackage{multirow}
\usepackage{booktabs}
\usepackage{array}
\usepackage{subcaption}
\usepackage{amssymb}
\usepackage{newfloat}
\usepackage{listings}
\DeclareCaptionStyle{ruled}{labelfont=normalfont,labelsep=colon,strut=off} 
\floatstyle{ruled}
\newfloat{listing}{tb}{lst}{}
\floatname{listing}{Listing}
\title{HeadHunt-VAD: Hunting Robust Anomaly-Sensitive Heads in MLLM for Tuning-Free Video Anomaly Detection}

\author{
    Zhaolin Cai\textsuperscript{\rm 1}, Fan Li\textsuperscript{\rm 2}\thanks{Corresponding Author}, Ziwei Zheng\textsuperscript{\rm 2}, Haixia Bi\textsuperscript{\rm 2}, Lijun He\textsuperscript{\rm 2}
}
\affiliations{
    \textsuperscript{\rm 1}School of Computer Science and Technology, Xinjiang University\\

    \textsuperscript{\rm 2}School of Information and Communications Engineering, Xi'an Jiaotong University\\
    107552301311@stu.xju.edu.cn, lifan@mail.xjtu.edu.cn,
    ziwei.zheng@stu.xjtu.edu.cn, \\haixia.bi@xjtu.edu.cn, lijunhe@mail.xjtu.edu.cn
}

\begin{document}

\maketitle
\begin{abstract}
Video Anomaly Detection (VAD) aims to locate events that deviate from normal patterns in videos. Traditional approaches often rely on extensive labeled data and incur high computational costs. Recent tuning-free methods based on Multimodal Large Language Models (MLLMs) offer a promising alternative by leveraging their rich world knowledge. However, these methods typically rely on textual outputs, which introduces information loss, exhibits normalcy bias, and suffers from prompt sensitivity, making them insufficient for capturing subtle anomalous cues. To address these constraints, we propose HeadHunt-VAD, a novel tuning-free VAD paradigm that bypasses textual generation by directly hunting robust anomaly-sensitive internal attention heads within the frozen MLLM. Central to our method is a Robust Head Identification module that systematically evaluates all attention heads using a multi-criteria analysis of saliency and stability, identifying a sparse subset of heads that are consistently discriminative across diverse prompts. Features from these expert heads are then fed into a lightweight anomaly scorer and a temporal locator, enabling efficient and accurate anomaly detection with interpretable outputs. Extensive experiments show that HeadHunt-VAD achieves state-of-the-art performance among tuning-free methods on two major VAD benchmarks while maintaining high efficiency, validating head-level probing in MLLMs as a powerful and practical solution for real-world anomaly detection. 
\end{abstract}

\begin{figure}[t]
    \centering
    \includegraphics[width=1\linewidth]{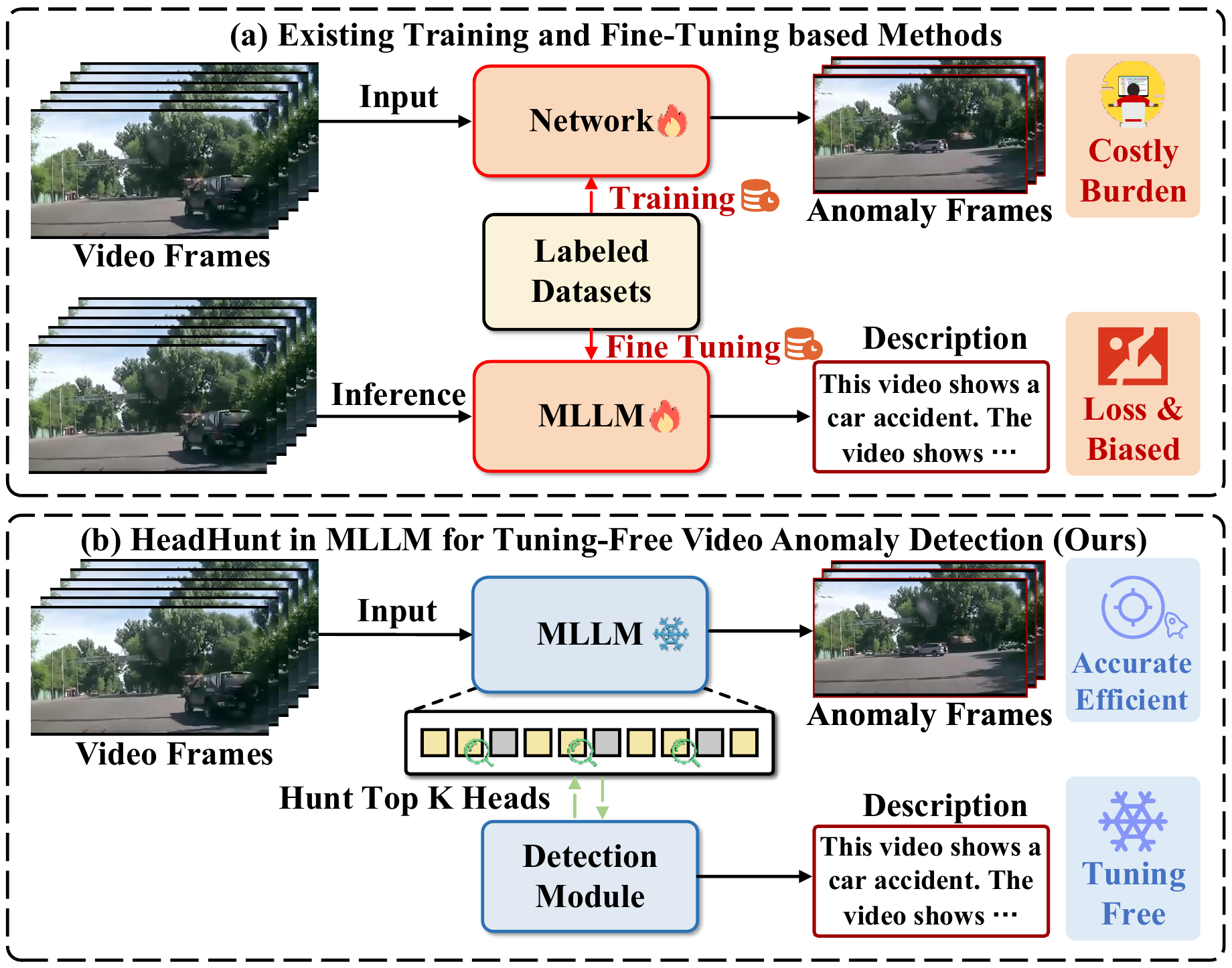}
    \caption{Comparison of VAD methods. Existing methods rely on training or fine tuning with large-scale datasets. HeadHunt-VAD is a tuning-free approach that detect anomalies using sparse expert heads within a frozen MLLM.}
    \label{fig:first}
    \vspace{-2mm}
\end{figure}

\section{Introduction}

Video Anomaly Detection (VAD) aims to identify and localize events that deviate from normal patterns in video sequences, which is a critical task for ensuring public safety \cite{sultani2018RealWorld}, industrial quality control \cite{roth2022Total}, and autonomous driving \cite{yao2023DoTA}. Traditional paradigms including unsupervised \cite{lv2021Learning}, weakly-supervised \cite{wu2024Weakly}, and fully-supervised  methods \cite{liu2018Future}, have made significant progress in VAD. However, these approaches often struggle to generalize to diverse anomalies and typically demand large-scale annotated training data along with substantial computational resources, hindering their scalability and practical deployment in real-world settings.

The emergence of Multimodal Large Language Models (MLLMs) \cite{liu2023Visual} has presented novel avenues for VAD, leveraging their powerful cross-modal reasoning and adaptability. While some methods adapt MLLMs to VAD through supervised fine-tuning \cite{zhang2024HolmesVAD}, this reintroduces the issues of data dependency and computational costs. Recent tuning-free VAD methods attempt to address these constraints \cite{zanella2024Harnessing} by prompting the model to reason and generate textual descriptions to detect anomalies. However, these methods mostly rely on the final textual output of MLLMs, which suffer from three key limitations. First, converting high-dimensional visual information into natural language inevitably incurs information loss, potentially omitting subtle yet crucial anomalous cues. \cite{zhang2025Crossmodal}. Second, MLLMs exhibit normalcy bias, tending to describe common objects while ignoring unusual details that define an anomaly. Third, their outputs are sensitive to prompt phrasing, often producing inconsistent predictions for semantically equivalent queries on the same video. These limitations motivate a shift away from textual outputs towards probing the internal representations of MLLMs.

Recent studies revealed that intermediate layers of large language models contain richer representations than output layer \cite{chen2024Bigger, sun2024Massive, skean2025Layer}. While some works have started to leverage features from entire intermediate layers for diverse tasks including VAD \cite{orgad2024LLMs,cai2025hiprobe}, this is still a coarse-grained approach. A transformer layer aggregates outputs from multiple attention heads \cite{vaswani2023Attention}, each with potential functional specializations \cite{zheng2025spotrisksspeakingunraveling}. The discriminative signals from a few heads risk being drowned out by outputs from heads focusing on mundane background features, therefore leading to representation dilution. Although prior work has analyzed heads for passive interpretation or guiding model pruning for efficiency \cite{baan2019Understanding, jin2024MoH}. The proactive identification and utilization of heads for VAD remains unexplored. This motivates a shift from layer-level to head-level representation analysis, aiming to identify and leverage fine-grained, informative attention heads for improved anomaly detection.

In this paper, we propose HeadHunt-VAD, a novel tuning-free paradigm for video anomaly detection that leverages sparse, expert attention heads within frozen Multimodal Large Language Models (MLLMs). As illustrated in Figure~\ref{fig:first}, our approach diverges from conventional methods by directly investigating internal representations of MLLM, mitigating the need for costly training and the problem of information loss. To overcome challenges of representation dilution, normalcy bias, and prompt sensitivity of MLLMs, we introduce the Robust Head Identification (RHI) module, which systematically hunts for a sparse set of anomaly-sensitive heads through multi-criteria analysis of saliency and stability across diverse prompts. Features from these expert heads are then used to construct a lightweight anomaly scorer and temporal locator via few-shot probing, requiring as little as 1\% of the training sets. This strategy effectively minimizing data dependency and supervision while enabling accurate anomaly detection and localization without any fine-tuning of the MLLM. We evaluate HeadHunt-VAD on two benchmark datasets including UCF-Crime \cite{sultani2018RealWorld} and XD-Violence \cite{wu2020not}, comprehensive experiments demonstrate the effectiveness of HeadHunt-VAD in video anomaly detection.

Our main contributions are summarized as follows:

\begin{itemize}
    \item We propose HeadHunt-VAD, a novel tuning-free paradigm for video anomaly detection that for the first time proactively hunting sparse, expert attention heads within a frozen MLLM, thereby addressing the issues of information loss and representation dilution.

    \item We introduce a novel Robust Head Identification (RHI) module, a prompt-invariant selection module that discovers anomaly-sensitive heads through saliency and stability analysis, effectively addressing the challenge of prompt sensitivity and enhancing robustness.
    
    \item We conduct extensive experiments on UCF-Crime and XD-Violence benchmarks, HeadHunt-VAD achieves state-of-the-art performance among tuning-free methods, achieved strong performance without any MLLM fine-tuning and with remarkable data efficiency.
\end{itemize}

\section{Related Work}
\subsection{Traditional Video Anomaly Detection}

Traditional VAD methods are typically categorized into supervised \cite{liu2018Future,landi2019Anomaly}, weakly-supervised \cite{li2022scale, wang2024Weakly, zhang2024GlanceVAD}, and unsupervised \cite{tur2023Unsupervised} paradigms. Supervised approaches achieve high accuracy by leveraging frame-level anomaly annotations but require costly and labor-intensive labeling. To reduce this burden, weakly supervised methods use video-level labels for training but struggle with subtle cues and may exhibit biased predictions. Unsupervised methods train exclusively on normal video data to model typical patterns and detect deviations \cite{hasan2016Learning, xu2017Detecting, yang2023Video}. Despite their effectiveness in specific settings, these methods are constrained by their reliance on large-scale training data, limiting their scalability and applicability in the real world.

\subsection{Video Anomaly Detection with LLMs and MLLMs}

The advent of large language models(LLM) \cite{touvron2023llamaopenefficientfoundation} and multimodal LLMs \cite{zhu2023MiniGPT4} has introduced new approaches to VAD, which can be broadly categorized into fine-tuning and tuning-free paradigms. Fine-tuning methods adapt pre-trained MLLMs for VAD \cite{zhang2023VideoLLaMA, yuan2024Surveillance, zhang2024HolmesVAU}, achieving strong results but reintroducing the need for extensive labeled data and computational resources. In contrast, tuning-free methods leverage the powerful cross-modal reasoning and zero-shot capabilities of frozen MLLMs for VAD \cite{shao2025EventVAD}. For example, LAVAD \cite{zanella2024Harnessing} detects anomalies by prompting MLLM to generate descriptions and reason with extra LLM; VERA \cite{ye2024VERA} employs verbalized learning to elicit more effective reasoning with MLLM through optimized prompts. However, these methods risk losing subtle visual patterns during the vision-to-text translation. This limitation motivates bypassing the textual output to probe more direct internal representations in MLLM.

\subsection{Internal Analysis in LLMs and MLLMs}

Recent research has demonstrated that intermediate layers of LLMs often contain richer representations than the final outputs \cite{jin2024Exploring, ju2024How, merullo2024Talking}, with mid-layer features shown to enhance performance in diverse downstream tasks \cite{skean2025Layer, orgad2024LLMs}. Recent work in MLLMs has revealed that the integration of visual and linguistic information occurs in the middle layers \cite{zhang2024Investigating, zhang2025Crossmodal}. Moreover, recent work has identified an information-rich phenomenon in intermediate layers, where internal features exhibit enhanced discriminative power for VAD \cite{cai2025hiprobe}. However, these analyses remain at the layer level. Although some studies has explored attention heads for model pruning \cite{baan2019Understanding, li2023Interpreting, jin2024MoH}, the proactive and goal-driven utilization of specific attention heads for video anomaly detection remains an underexplored and promising frontier.

\section{Methodology}

\subsection{Preliminaries: Representation Dilution in MLLMs}
\label{sec:preliminaries}

\begin{figure}[t]
    \centering
    \begin{subfigure}[b]{0.49\columnwidth}
        \includegraphics[width=\textwidth]{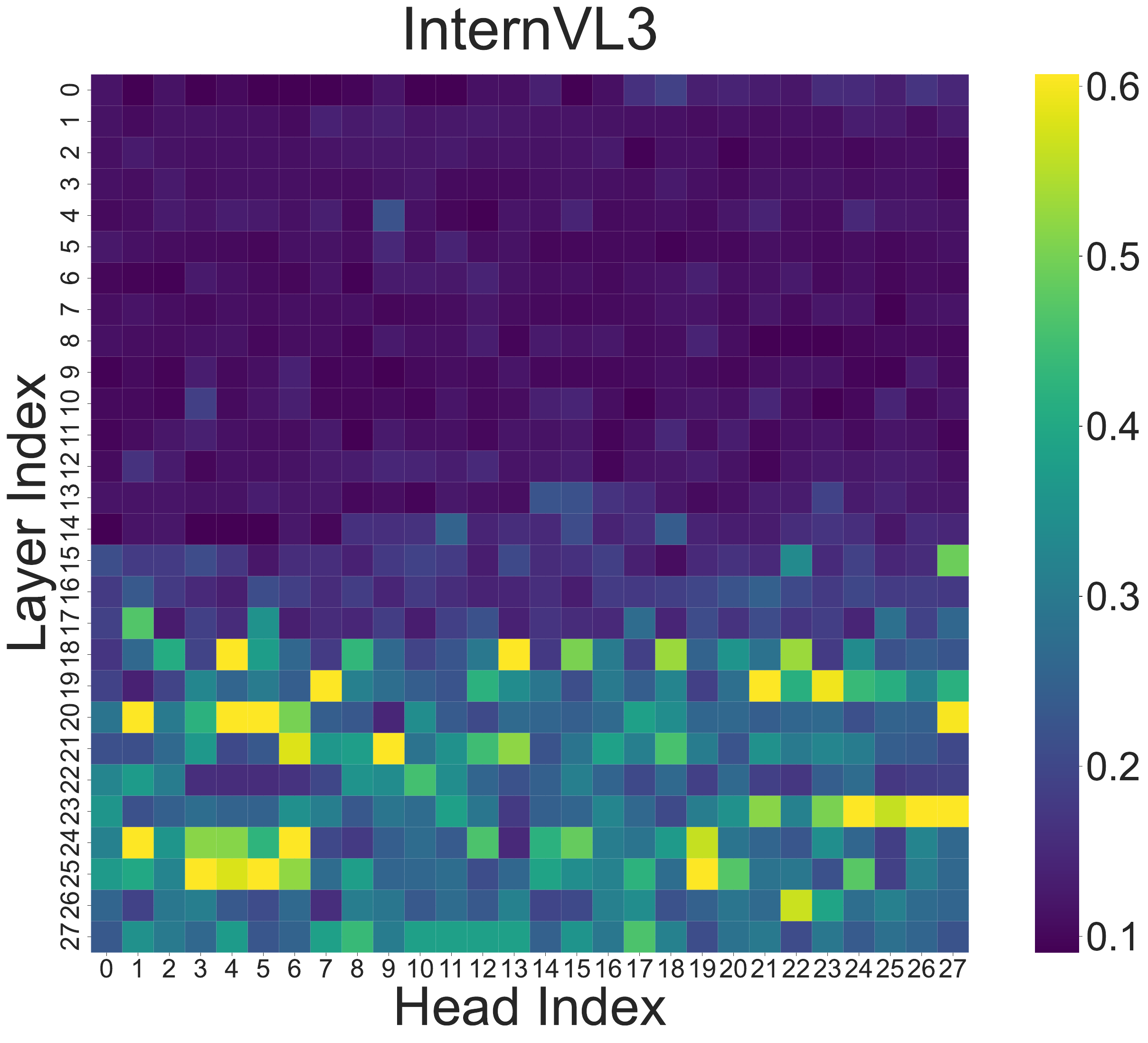}
        \label{fig:internvl3}
    \end{subfigure}
    \begin{subfigure}[b]{0.49\columnwidth}
        \includegraphics[width=\textwidth]{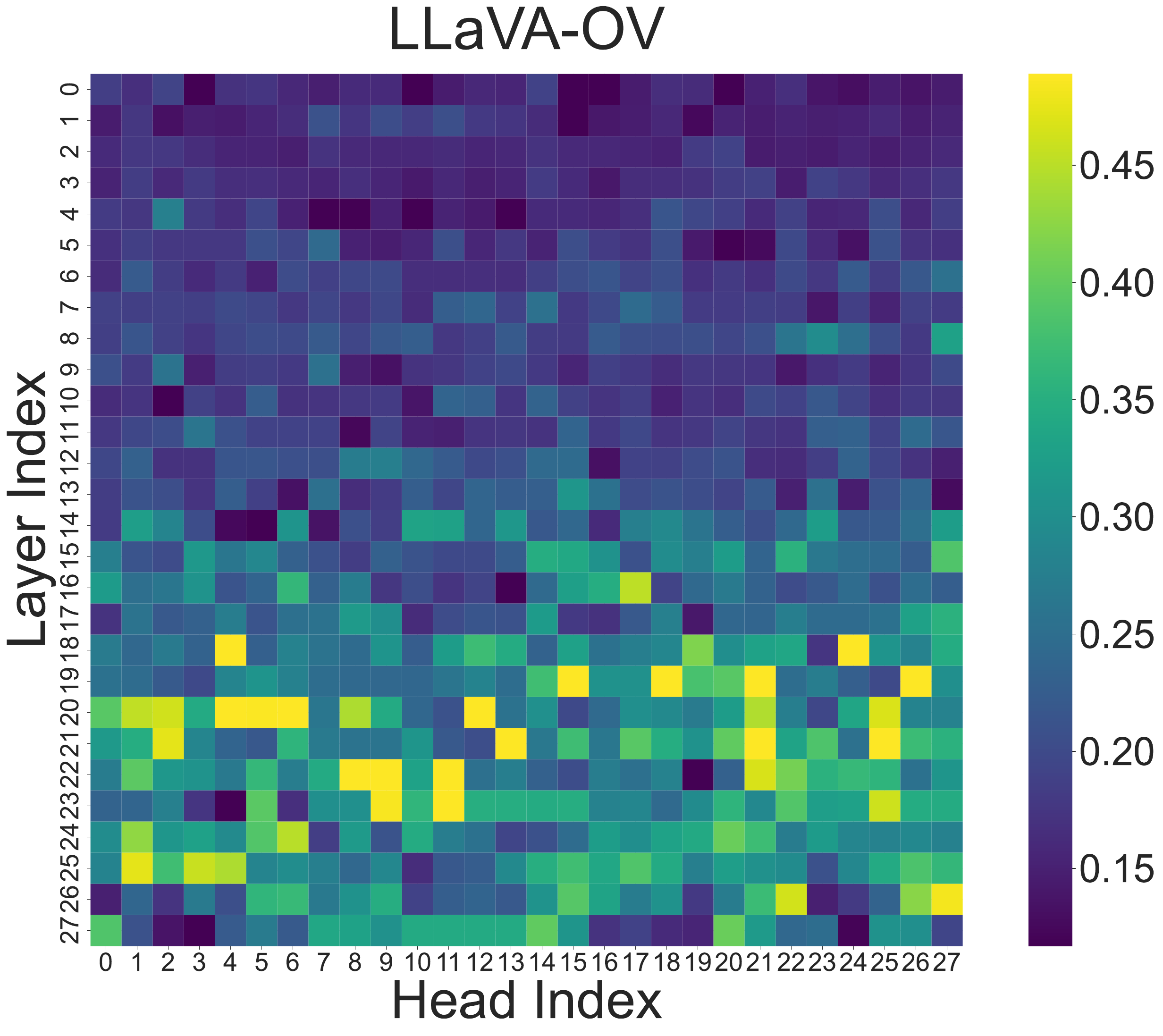}
        \label{fig:llavaov}
    \end{subfigure}
    \caption{Visualization of attention head heatmaps in MLLMs including InternVL3 and LLaVA-OV.}
    \label{fig:heatmap}
\end{figure}

The multi-head attention (MHA) module is the cornerstone of the transformer architecture, which underpins Multimodal Large Language Models (MLLMs). Given an input feature sequence $\mathbf{X} \in \mathbb{R}^{L \times D}$, where $L$ is the sequence length and $D$ is the model dimension, MHA projects the input into queries ($\mathbf{Q}$), keys ($\mathbf{K}$), and values ($\mathbf{V}$). The output of the $j$-th attention head, $\mathbf{h}_j \in \mathbb{R}^{L \times d_h}$, is computed as:
\begin{equation}
\mathbf{h}_j = \text{Attention}(\mathbf{Q}\mathbf{W}_j^Q, \mathbf{K}\mathbf{W}_j^K, \mathbf{V}\mathbf{W}_j^V)
\end{equation}
where $\mathbf{W}_j^Q, \mathbf{W}_j^K, \mathbf{W}_j^V \in \mathbb{R}^{D \times d_h}$ are the projection matrices for the $j$-th head, and $d_h$ is the dimension of head. The outputs of all $N_h$ heads are then concatenated and projected by a final linear layer $\mathbf{W}^O \in \mathbb{R}^{N_h d_h \times D}$ to produce the aggregated output:
\begin{equation}
\mathbf{X}' = \text{Concat}(\mathbf{h}_1, \dots, \mathbf{h}_{N_h})\mathbf{W}^O
\end{equation}
However, this final aggregation step can be detrimental for fine-grained tasks. It averages outputs from all heads, causing sharp, discriminative signals from specialized, anomaly-sensitive heads to be diluted by less relevant ones. Therefore we term this issue as representation dilution. As illustrated in Figure~\ref{fig:heatmap}, individual heads in intermediate layers often exhibit superior discriminative power compared to the aggregated final output, where crucial signals are lost. This observation motivates our work, as any method relying on the final, diluted output $\mathbf{X}'$ is inherently handicapped for tasks like video anomaly detection that demand high sensitivity to subtle cues. Therefore, HeadHunt-VAD is designed to circumvent this issue by operating directly on the pre-aggregation outputs from all heads across all layers. We denote the set of all head outputs as $\{\mathbf{h}_k\}_{k=1}^{N_{\text{total}}}$, where $k$ is a global index for each head and $N_{\text{total}} = N_{\text{layers}} \times N_h$ is the total number of heads in the MLLM. This allows us to directly identify and exploit the most informative heads.

\subsection{HeadHunt-VAD Framework Overview}
\label{sec:framework_overview}

\begin{figure*}[t]
\centering
\includegraphics[width=1\textwidth]{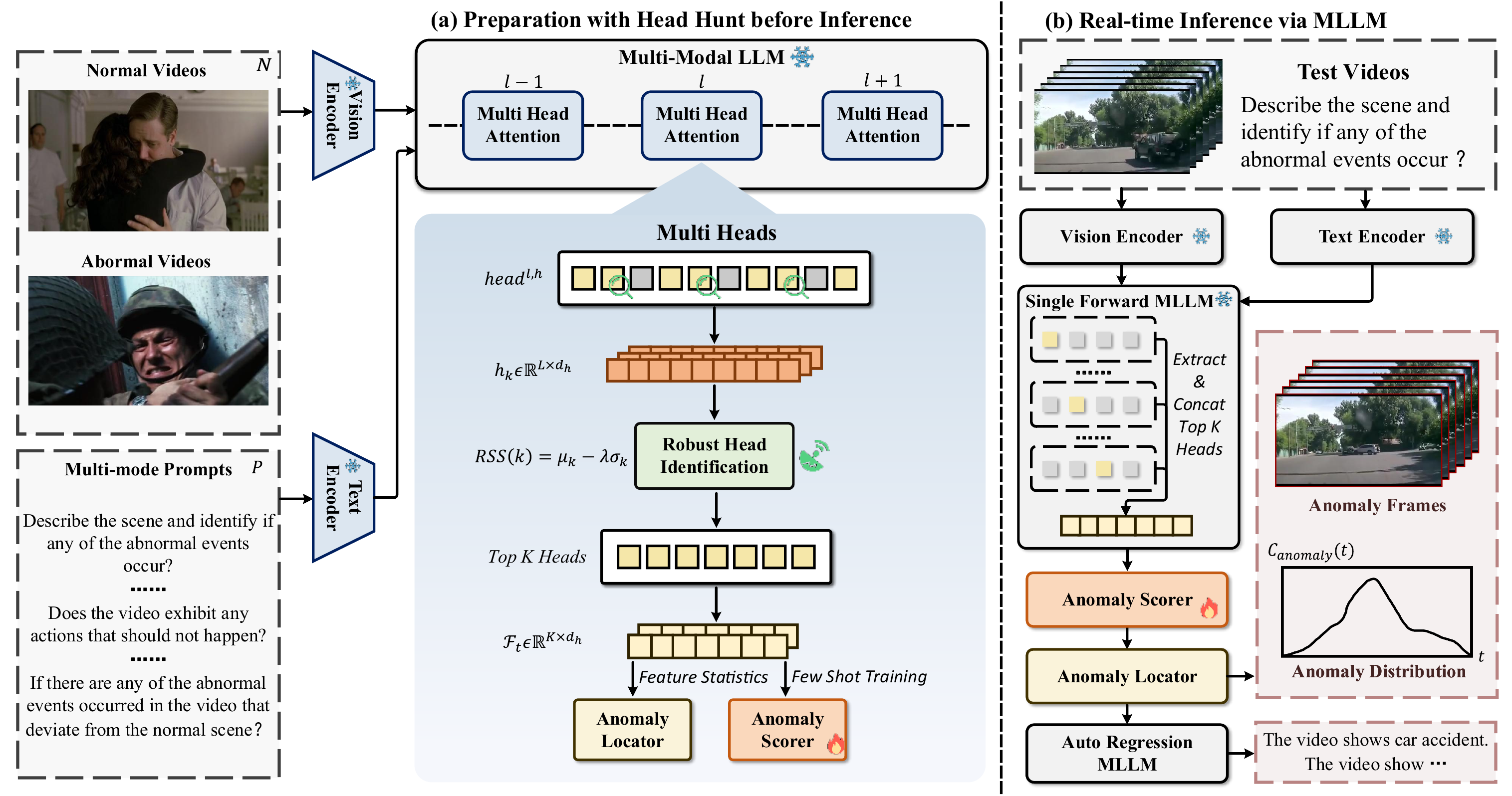}
\caption{The overall architecture of HeadHunt-VAD. The offline HeadHunt phase identifies consensus expert heads and prepares downstream modules. The efficient online inference phase leverages these for real-time detection.}
\label{fig:framework}
\end{figure*}

HeadHunt-VAD is a highly efficient paradigm for video anomaly detection by leveraging a frozen Multimodal Large Language Model (MLLM) without any fine-tuning. As depicted in Figure~\ref{fig:framework}, our approach is structured into two main stages: an offline preparation stage and a real-time online inference stage. In the offline stage, the core HeadHunt process commences with the Robust Head Identification (RHI) module, which systematically evaluates all attention heads in multi-criteria analysis to select a prompt-robust set of consensus expert heads based on their discriminative power. Based on features from selected subset of heads, we construct two lightweight modules: a logistic regression-based Anomaly Scorer and a calibrated Temporal Locator. The online stage is engineered for maximum throughput. For an incoming video, it performs a single forward pass to perform targeted feature extraction from the identified expert heads. These features are then processed by the lightweight scorer and locator for precise, real-time anomaly localization and finally generate descriptions for detected anomalies.

\subsection{Offline Stage: Robust Head Identification and Module Preparation}
\label{sec:offline_stage}
The offline stage focus on head identification and module construction to enable efficient and accurate real-time inference in the subsequent online stage. All attention heads across layers are treated as a unified set, indexed by $k \in \{1, \dots, N_{\text{total}}\}$, where $N_{\text{total}} = N_{\text{layers}} \times N_h$.

\subsubsection{Head Saliency Characterization}
\label{sec:saliency_char}

To evaluate the usefulness of each head, we first extract a representative feature vector from MLLM. For a given video and text prompt, we perform a single forward pass through the frozen MLLM and intercept the output of each head $\mathbf{h}_k \in \mathbb{R}^{L \times d_h}$ before the final aggregation layer. From this matrix, we extract the feature vector $\mathbf{x}_k \in \mathbb{R}^{d_h}$ corresponding to the first generated token, as it encapsulates a summary of the input sequence for the initial generative decision and avoids the cost of the full auto-regressive decoding process. By processing all videos in our calibration set, we form feature sets $\mathcal{X}_{k,n}$ (normal) and $\mathcal{X}_{k,a}$ (abnormal) for each head $k$. To obtain a holistic and robust assessment of discriminative power with each head after the extraction, we construct comprehensive metrics to effectively calculate the saliency score. The multi-faceted evaluation spans across statistical, geometric, and information-theoretic perspectives, thereby ensuring our selection is not biased by a single criterion. The saliency of head $k$ under a given prompt $p_m$ is then characterized as follows. Detailed derivations and pipeline algorithms are provided in the supplementary materials.

\begin{itemize}
    \item \textbf{Linear Discriminant Analysis Score ($S_{\text{LDA}}$):} Measures linear separability, aligning with our goal of finding features suitable for an efficient linear classifier. A higher score signifies greater class separation.
    
    \begin{equation}
    S_{\text{LDA}}(k) = (\boldsymbol{\mu}_{k,a} - \boldsymbol{\mu}_{k,n})^T (\mathbf{S}_{W,k})^{-1} (\boldsymbol{\mu}_{k,a} - \boldsymbol{\mu}_{k,n})
    \end{equation}

    where $\boldsymbol{\mu}_{k,c}$ and $\mathbf{S}_{W,k}$ are the class-specific mean and within-class scatter matrix for head $k$.                                                                 ,

    \item \textbf{Symmetrized KL Divergence ($S_{\text{KL}}$):} Quantifies the dissimilarity between the probability distributions of normal and abnormal representations, which are modeled as multivariate Gaussian distributions, thereby capturing non-linear statistical differences.
    {
    \small
    \begin{equation}
    S_{\text{KL}}(k) = \frac{1}{2} [D_{KL}(\mathcal{N}_{k,a} || \mathcal{N}_{k,n}) + D_{KL}(\mathcal{N}_{k,n} || \mathcal{N}_{k,a})]
    \end{equation}
    }
    
    \item \textbf{Maximum Mean Discrepancy ($S_{\text{MMD}}$):} Measures the distributional discrepancy in a Reproducing Kernel Hilbert Space (RKHS), making it sensitive to complex differences in distribution shape.
    {\small
    \begin{equation}
    S_{\text{MMD}}(k)^2 = \left\| \frac{1}{|\mathcal{X}_{k,a}|} \sum_{\mathbf{x} \in \mathcal{X}_{k,a}} \phi(\mathbf{x}) - \frac{1}{|\mathcal{X}_{k,n}|} \sum_{\mathbf{y} \in \mathcal{X}_{k,n}} \phi(\mathbf{y}) \right\|_{\mathcal{H}}^2
    \end{equation}
    }

    where $\phi(\cdot)$ is the kernel-induced feature map into the RKHS $\mathcal{H}$.
    
    \item \textbf{Normalized Mutual Information ($S_{\text{NMI}}$):} Measures the statistical dependency between feature representations and ground-truth labels. We perform K-Means clustering ($K=2$) on the features to get predicted labels $Y_{\text{pred}}^{(k)}$.
    \begin{equation}
    S_{\text{NMI}}(k) = \frac{I(Y_{\text{true}}, Y_{\text{pred}}^{(k)})}{\sqrt{H(Y_{\text{true}})H(Y_{\text{pred}}^{(k)})}}
    \end{equation}
    where $I(\cdot, \cdot)$ is the mutual information and $H(\cdot)$ is the entropy.

\end{itemize}
Although these metrics provide a comprehensive evaluation of discriminative capability, identifying universally effective heads requires assessing their stability across diverse textual inputs. We therefore introduce the Robust Head Selection.

\subsubsection{Robust Head Selection}
\label{sec:robust_selection}
To identify heads that are universally effective, we assess their stability across diverse textual inputs. We evaluate each head over a set of $M$ diverse prompts, $\mathcal{P} = \{p_1, \dots, p_M\}$. For each head $k$ and prompt $p_m$, we first compute every saliency scores and normalize them to $[0, 1]$ across all heads. The score $S(k, p_m)$ is the average of normalized saliency scores.

The mean performance $\mu_k$ and instability (standard deviation) $\sigma_k$ are calculated for each head across the prompt set:
\begin{equation}
  \mu_k = \frac{1}{M} \sum_{m=1}^{M} S(k, p_m)
\end{equation}
\begin{equation}
  \sigma_k = \sqrt{\frac{1}{M} \sum_{m=1}^{M} \left(S(k, p_m) - \mu_k\right)^2}
\end{equation}
Inspired by risk-aversion principles, we define a Robust Saliency Score (RSS) to favor heads with both high average performance and low variance across prompts:
\begin{equation}
\text{RSS}(k) = \mu_k - \lambda \sigma_k
\end{equation}

where $\lambda$ is a hyperparameter that controls the trade-off between performance and stability. A head that performs well for one prompt but poorly for another will have a high $\sigma_k$ and thus be penalized, as we seek heads that are not only high returns (high mean saliency $\mu_k$) but also low risk (low performance volatility $\sigma_k$ across prompts). We calculate and rank all heads by their RSS and select the top-$K$ heads, whose indices form our consensus expert head set $\mathcal{I}^*$.

\subsubsection{Anomaly Scorer Training} 
The anomaly scorer is implemented using logistic regression, chosen for its efficiency and interpretability. For each video $i$ in the calibration set, a composite feature vector $\mathbf{z}_i \in \mathbb{R}^{K \cdot d_h}$ is constructed by concatenating feature $\{\mathbf{x}_k\}_{k \in \mathcal{I}^*}$ from the $K$ expert heads. Given the training set $\{(\mathbf{z}_i, y_i^{\text{(v)}})\}_{i=1}^{N}$, where $y_i^{\text{(v)}} \in \{0, 1\}$ is the video-level label, the model learns a weight vector $\mathbf{w}$ and bias $b$ to predict the anomaly probability:
\begin{equation}
p_i^{\text{(v)}} = \sigma(\mathbf{w}^T \mathbf{z}_i + b)
\end{equation}

where $\sigma(\cdot)$ is the sigmoid function. The model is trained by minimizing the binary cross-entropy loss:
{
\small
\begin{equation}
\mathcal{L} = -\frac{1}{N} \sum_{i=1}^N \left[ y_i^{\text{(v)}} \log p_i^{\text{(v)}} + (1 - y_i^{\text{(v)}}) \log(1 - p_i^{\text{(v)}}) \right]
\end{equation}
}
This lightweight model effectively leverages the discriminative features from the expert heads while ensuring computational efficiency and accuracy.

\subsubsection{Temporal Locator Calibration}
To convert anomaly probabilities into coherent temporal event predictions, a temporal locator is introduced. This component operates on a sequence of probabilities $\{p_t\}$, which are generated by applying the trained Anomaly Scorer to the feature vector of each individual temporal segment from the validation set. The locator then refines the resulting raw probability sequence $\mathbf{p}$ by applying a transformation that includes convolution with a 1D Gaussian kernel, $G_{\sigma_g}$, followed by binarization using a threshold $\tau$.

The standard deviation $\sigma_g$ of the Gaussian kernel and the threshold $\tau$ are determined by optimizing the frame-level F1-score on a validation set via grid search. For each candidate pair $(\sigma_g, \tau)$, the validation videos are processed by first generating the raw probability sequence $\mathbf{p}$ and then applying the complete transformation:
\begin{equation}
p'_t = (\mathbf{p} * G_{\sigma_g})_t = \sum_{j} p_j \cdot G(t-j; \sigma_g)
\end{equation}

The resulting binarized outputs are compared with ground-truth annotations to compute the F1-score. The parameter combination $(\sigma_g^*, \tau^*)$ that maximizes the score is selected. This data-driven calibration ensures the Temporal Locator is configured for precise anomaly localization.

\subsection{Online Stage: Real-time Video Anomaly Detection}
\label{sec:online_stage}
The online stage with MLLMs focuses on processing unseen videos to detect and localize anomaly frames and optionally provide comprehensive anomaly descriptions.

\subsubsection{Single Forward Pass with Feature Extraction}
An incoming video is first divided into a sequence of non-overlapping temporal segments $\{S_1, S_2, \dots, S_T\}$. For each segment $S_t$, we uniformly sample $F$ frames. These frames, along with a task-specific textual prompt, are processed in a single forward pass through the frozen MLLM. During this pass, we perform targeted feature extraction by retrieving and concatenating the outputs from the $K$ expert heads in our consensus set $\mathcal{I}^*$, thereby avoiding the costly auto-regressive decoding. Following the procedure from the offline stage, we extract the feature corresponding to the first token from each selected head and concatenate them to form a single, robust feature vector $\mathbf{f}_t \in \mathbb{R}^{K \cdot d_h}$ for segment $S_t$.

\subsubsection{Frame-level Anomaly Scoring and Localization}
The feature vector $\mathbf{f}_t$ is fed into the pre-trained Anomaly Scorer to compute the anomaly probability for each segment:
\begin{equation}
p_t = \sigma(\mathbf{w}^T \mathbf{f}_t + b)
\end{equation}
This yields a sequence of anomaly scores $\mathbf{p} = (p_1, \dots, p_T)$ for the entire video. The probability sequence $\mathbf{p}$ is then processed by the calibrated Temporal Locator. The sequence is first smoothed using the pre-calibrated Gaussian kernel $G_{\sigma_g^*}$ to enforce temporal consistency, producing a smoothed sequence $\mathbf{p}'$. The final detection $\hat{y}_t \in \{0, 1\}$ is made by applying the calibrated threshold $\tau^*$:
\begin{equation}
\hat{y}_t = [p'_t > \tau^*]
\end{equation}
where $[\cdot]$ denotes the Iverson bracket. Consecutive segments where $\hat{y}_t=1$ are grouped to yield precise temporal localizations of anomalous events.

\subsubsection{Event-level Explanation Generation}
For enhanced interpretability, detected anomalous clips in a video can be passed back to the full auto-regressive MLLM. Given the clips and textual inputs, the model generates a natural language explanation of the event. This final step completes the process from detection to understanding, providing a comprehensive and interpretable explanations for videos.

\section{Experiments}

\subsection{Experimental Settings}

\subsubsection{Datasets and evaluation metrics}
We evaluate our method on two widely used benchmarks for video anomaly detection: UCF-Crime \cite{sultani2018RealWorld} and XD-Violence \cite{wu2020not}. These datasets contain long, untrimmed videos with real-world anomalies, providing a realistic and challenging evaluation environment.

For performance evaluation, we use standard metrics from the literature: frame-level AUC (Area Under the ROC Curve) for UCF-Crime and average precision (AP) for XD-Violence. Both metrics are widely used in prior work and provide a comprehensive assessment of detection performance, higher values indicate better results.

\subsubsection{Implementation Details}
We use the InternVL3 model as the frozen MLLM backbone of HeadHunt-VAD. Each video is segmented at 48 frames intervals and uniformly sample $F=16$ frames as the input to MLLM. The offline phase use a small calibration subset comprising 1\% of the training data from each benchmark. The prompt set $\mathcal{P}$ consists of five varied prompts. We select the top $K=5$ expert heads for feature extraction and set the instability penalty to $\lambda=0.5$. The Anomaly Scorer is a logistic regression module. The Temporal Locator uses 1D Gaussian smoothing kernel $\sigma_g=1.5$ and threshold of $\tau=0.65$, both optimized for frame-level F1-score on the validation set. All experiments are conducted on a single NVIDIA RTX 4090 GPU. Detailed hyperparameter analysis, prompts, and further analysis are provided in the supplementary materials.

\begin{table}[t]
\centering
\small
\begin{tabular}{
@{}>{\centering\arraybackslash}c@{\hspace{4pt}} 
| @{}>{\hspace{4pt}}l@{} 
@{}c@{} 
@{\hspace{2pt}}c@{\hspace{2pt}} 
}
\toprule
\textbf{Mode} & \textbf{Methods} & \textbf{Backbone} & \textbf{AUC (\%)} \\
\midrule
\multirow{10}{*}{\centering\shortstack{Weakly \\ Supervised}} 
 & Wu et al. \shortcite{wu2020not} & I3D & 82.44 \\
 & MIST\shortcite{feng2021MIST} & I3D & 82.30 \\
 & RTFM\shortcite{tian2021Weaklysupervised} & I3D & 84.30 \\
 & S3R\shortcite{wu2022self} & I3D & 85.99 \\
 & MSL\shortcite{li2022SelfTraining} & I3D & 85.30 \\
 & UR-DMU\shortcite{zhou2023Dual} & I3D & 86.97 \\
 & MFGN\shortcite{chen2022MGFN} & I3D & 86.98 \\
 & Wu et al.\shortcite{wu2024Openvocabulary} & ViT & 86.40 \\
 & CLIP-TSA\shortcite{joo2023CLIPTSA} & ViT & 87.58 \\
 & Yang et al.\shortcite{yang2024Text} & ViT & 87.79 \\
 & VadCLIP\shortcite{wu2023VadCLIP} & ViT & 88.02 \\
\midrule
\multirow{3}{*}{\centering\shortstack{Self \\ Supervised}} 
 & TUR et al.\shortcite{tur2023Unsupervised} & Resnet & 66.85 \\
 & BODS\shortcite{wang2019GODS} & I3D & 68.26 \\
 & GODS\shortcite{wang2019GODS} & I3D & 70.46 \\
\midrule
\multirow{2}{*}{\centering\shortstack{Unsupervised}} 
 & GCL\shortcite{zaheer2022Generative} & ResNext & 71.04 \\
 & DYANNET\shortcite{thakare2023DyAnNet} & I3D & 84.50 \\
\midrule
\multirow{9}{*}{\centering\shortstack{Tuning-Free \\ Multimodal \\ VAD}} 
 & Zero-Shot CLIP\shortcite{radford2021Learning} & ViT & 53.16 \\
 & ZS ImageBind (Video)\shortcite{girdhar2023ImageBind} & ViT & 55.78 \\
 & ZS ImageBind (Image)\shortcite{girdhar2023ImageBind} & ViT & 53.65 \\
 & LLAVA-1.5\shortcite{liu2024Improved} & ViT & 72.84 \\
 & LAVAD\shortcite{zanella2024Harnessing} & ViT & 80.28 \\
 & EventVAD\shortcite{shao2025EventVAD} & ViT & 82.03 \\
 & VERA\shortcite{ye2024VERA} & ViT & 86.55 \\
 & HiProbeVAD\shortcite{cai2025hiprobe} & ViT & 86.72 \\
 & \textbf{HeadHunt-VAD} & \textbf{ViT} & \textbf{87.03} \\
\bottomrule
\end{tabular}
\caption{Comparison with existing methods on the UCF-Crime dataset.}
\label{tab:comparison_ucf}
\end{table}

\begin{table}[t]
\centering
\small
\begin{tabular}{
@{}>{\centering\arraybackslash}c@{\hspace{4pt}} 
| @{}>{\hspace{4pt}}l@{} 
@{}c@{} 
@{\hspace{2pt}}c@{\hspace{2pt}} 
}
\toprule
\textbf{Mode} & \textbf{Methods} & \textbf{Backbone} & \textbf{AP (\%)} \\
\midrule
\multirow{10}{*}{\centering\shortstack{Weakly \\ Supervised}} 
 & Wu et al.\shortcite{wu2020not} & I3D & 73.20 \\
 & RTFM\shortcite{tian2021Weaklysupervised} & I3D & 77.81 \\
 & MSL\shortcite{li2022SelfTraining} & I3D & 78.28 \\
 & MFGN\shortcite{chen2022MGFN} & I3D & 79.19 \\
 & S3R\shortcite{wu2022self} & I3D & 80.26 \\
 & UR-DMU\shortcite{zhou2023Dual} & I3D & 81.66 \\
 & Wu et al.\shortcite{wu2024Openvocabulary} & ViT & 66.53 \\ 
 & CLIP-TSA\shortcite{joo2023CLIPTSA} & ViT & 82.19 \\
 & Yang et al.\shortcite{yang2024Text} & ViT & 83.68 \\
 & VadCLIP\shortcite{wu2023VadCLIP} & ViT & 84.51 \\
\midrule
\multirow{8}{*}{\centering\shortstack{Tuning-Free \\ Multimodal \\ VAD}} 
 & Zero-Shot CLIP\shortcite{radford2021Learning} & ViT & 17.83 \\
 & ZS ImageBind (Video)\shortcite{girdhar2023ImageBind}& ViT & 25.36 \\
 & ZS ImageBind (Image)\shortcite{girdhar2023ImageBind}& ViT & 27.25 \\
 & LLAVA-1.5\shortcite{liu2024Improved} & ViT & 50.26 \\
 & LAVAD\shortcite{zanella2024Harnessing} & ViT & 62.01 \\
 & EventVAD\shortcite{shao2025EventVAD} & ViT & 64.04 \\
 & HiProbeVAD\shortcite{cai2025hiprobe} & ViT & 82.15 \\
 & \textbf{HeadHunt-VAD} & \textbf{ViT} & \textbf{82.63} \\
\bottomrule
\end{tabular}
\caption{Comparison of existing methods on the XD-Violence dataset.}
\label{tab:comparison_xd}
\end{table}

\subsection{Comparison with State-of-the-Art Methods}
Tables~\ref{tab:comparison_ucf} and \ref{tab:comparison_xd} present the main results on the UCF-Crime and XD-Violence datasets. On UCF-Crime, HeadHunt-VAD achieves an AUC of 87.03\%, setting a new state-of-the-art among tuning-free and unsupervised methods. Our method also remains competitive with many weakly-supervised methods that require extensive training on large-scale labeled datasets. On the XD-Violence dataset, HeadHunt-VAD achieves an AP of 82.63\%, which represents an improvement over other tuning-free approaches and significantly narrows the performance gap to leading weakly-supervised models. The consistent and robust performance across these two benchmarks provides strong evidence for the effectiveness of our proposed approach for VAD.

\begin{figure}[t]
    \centering
    \begin{subfigure}[b]{0.45\columnwidth}
        \includegraphics[width=\textwidth]{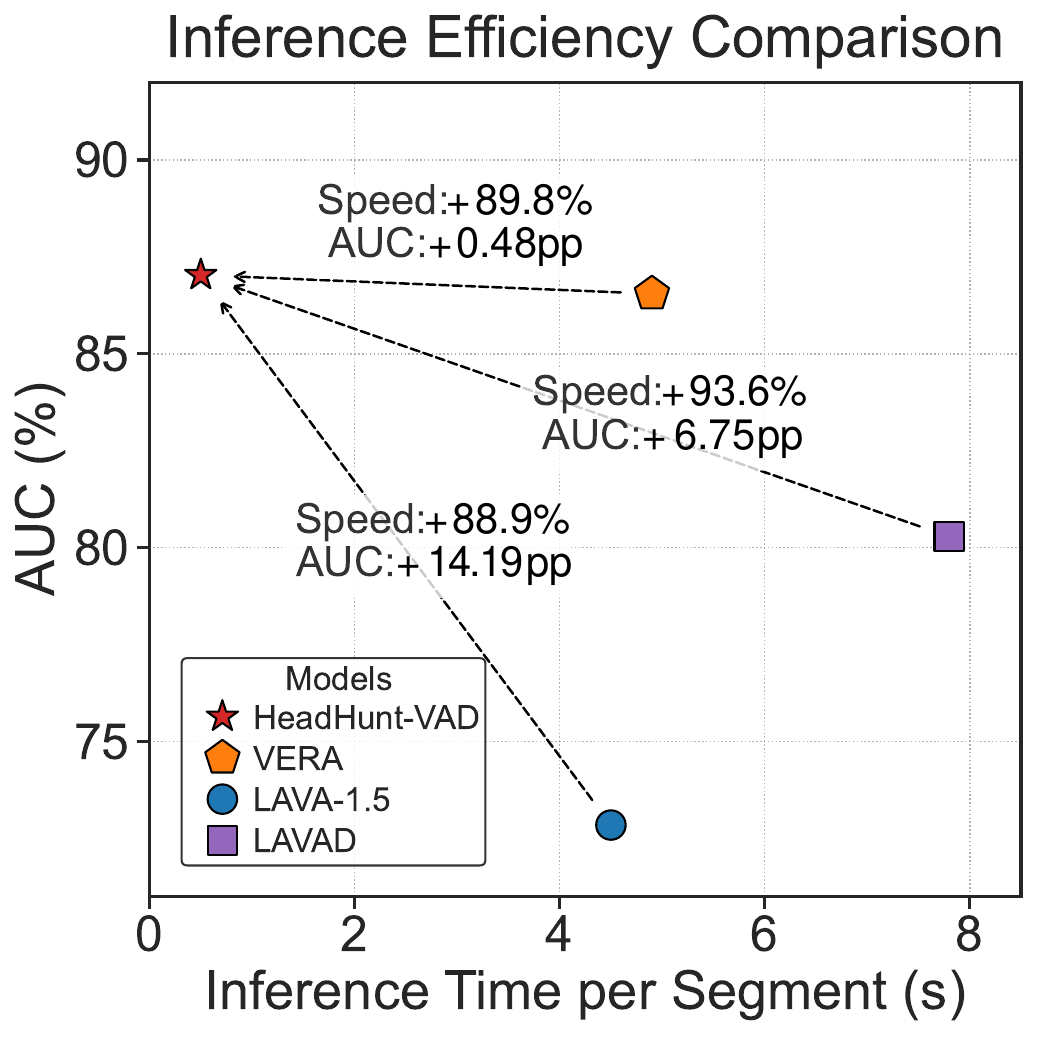}
        \label{fig:time}
    \end{subfigure}
    \begin{subfigure}[b]{0.54\columnwidth}
        \includegraphics[width=\textwidth]{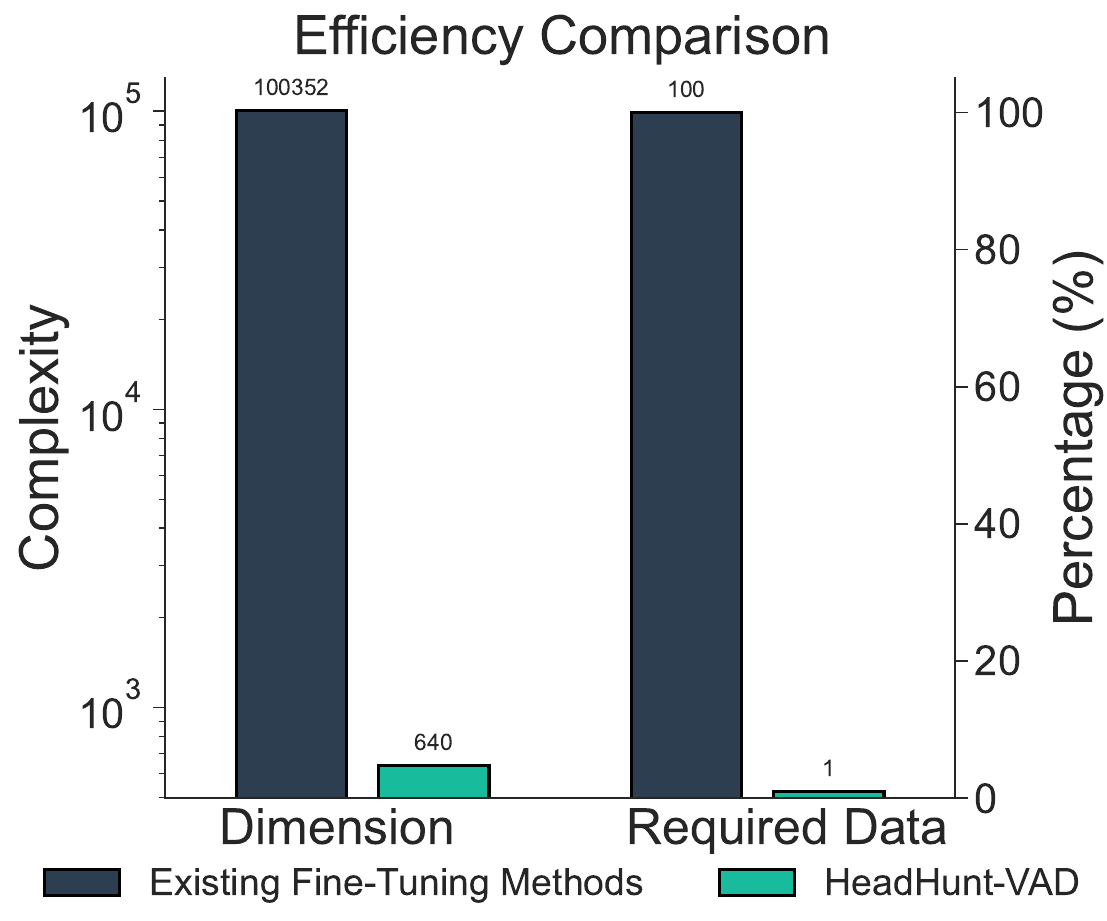}
        \label{fig:data&dimension}
    \end{subfigure}
    \caption{Efficiency comparison of HeadHunt-VAD including inference time, dimension complexity, and required data.}
    \label{fig:efficiency_analysis}
\end{figure}

\subsection{Efficiency Analysis}
HeadHunt-VAD achieves practical efficiency by reducing computational cost, model complexity, and data dependency. As shown in Figure~\ref{fig:efficiency_analysis}, our single forward pass strategy avoids costly auto-regressive decoding and yields a significant computational efficiency. This efficiency extends to scalability. Our few-shot calibration uses less than 1\% of the data required for fine-tuning approaches. By extracting features from only top-$K$ expert heads instead of all heads, we dramatically reduce the feature dimension from over 100K to just 640. The combination of fast detection, minimal data requirements, and low feature complexity validates HeadHunt-VAD as an effective and practical VAD solution.


\subsection{Ablation Studies}
\label{sec:ablation}

\begin{table}[t]
\centering
\begin{tabular}{lcc}
\toprule
\textbf{Method / Variation} & \textbf{AUC (\%)} & \textbf{AP (\%)} \\
\midrule
\textbf{HeadHunt-VAD (Full Model)} & \textbf{87.03} & \textbf{82.63} \\
\midrule
\multicolumn{3}{l}{\textit{Ablation on Robust Head Identification (RHI)}} \\
\cmidrule(r){1-3}
\quad w/ Full Layer Features & 80.15 & 72.10 \\
\quad w/ Random-K Heads & 66.65 & 45.33 \\
\quad w/ Single Coarse Prompt & 81.86 & 74.52 \\
\quad w/ Oracle (Detailed) Prompt & 87.11 & 82.95 \\
\midrule
\multicolumn{3}{l}{\textit{Ablation on Anomaly Scorer (Ours: Logistic Regression)}} \\
\cmidrule(r){1-3}
\quad SVM (Linear Kernel) & 84.95 & 76.52 \\
\quad MLP (2-Layer) & 87.25 & 82.81 \\
\midrule
\multicolumn{3}{l}{\textit{Ablation on Temporal Locator (Ours: $\tau=0.65$)}} \\
\cmidrule(r){1-3}
\quad w/o Gaussian Smoothing & 82.44 & 75.88 \\
\quad w/ Fixed Threshold ($\tau=0.25$) & 70.91 & 55.21 \\
\quad w/ Fixed Threshold ($\tau=0.50$) & 80.32 & 71.49 \\
\quad w/ Fixed Threshold ($\tau=0.75$) & 71.15 & 58.10 \\
\bottomrule
\end{tabular}
\caption{Effectiveness of the Robust Head Identification module, Anomaly Scorer, and the Temporal Locator.}
\label{tab:ablation_study}
\end{table}

\subsubsection{Effectiveness of the Robust Head Identification}
We evaluate the proposed Robust Head Identification (RHI) by comparing it with different feature extraction strategies (Table~\ref{tab:ablation_study}). Using all attention heads from the final layer leads to substantial performance degradation, while random head selection further deteriorates results, confirming the necessity of informed head selection. We also assess prompt sensitivity. A single generic prompt underperforms, and although a manually crafted oracle prompt yields slightly better results, it requires fixed prompt for all time. In contrast, RHI automatically identifies a robust subset across prompts, achieving comparable performance with different text inputs.

\subsubsection{Impact of Anomaly Scorer}
We analyze the role of the final anomaly scorer with other common lightweight models, with results presented in Table~\ref{tab:ablation_study}. The support vector machine (SVM) achieves performance comparable to the proposed method, while multi-layer perceptron (MLP) yields a marginal improvement of 0.22\% in AUC and 0.18\% in AP. We adopt logistic regression for its optimal balance of accuracy, computational efficiency, and interpretability, which aligns with overall emphasis on practicality and ease of deployment of our method.

\subsubsection{Impact of Temporal Locator}
Our temporal locator consists of two key components: temporal score smoothing and a data-driven threshold calibration, with results in Table~\ref{tab:ablation_study}. Removing the Gaussian smoothing step results in a significant AUC drop of 4.59\% on UCF-Crime, which demonstrates its crucial role in stabilizing frame-level predictions and reducing spurious noise from isolated, high-scoring frames. Furthermore, replacing our data-driven threshold calibration with fixed, arbitrary values leads to severe performance degradation. This confirms that a calibrated, data-aware threshold is indispensable for accurately segmenting anomalous events from raw scores.

\subsection{Qualitative Analyses}

\subsubsection{Feature Space Visualization}
\begin{figure}[t!]
    \centering
    \begin{subfigure}[b]{0.49\columnwidth}
        \includegraphics[width=\textwidth]{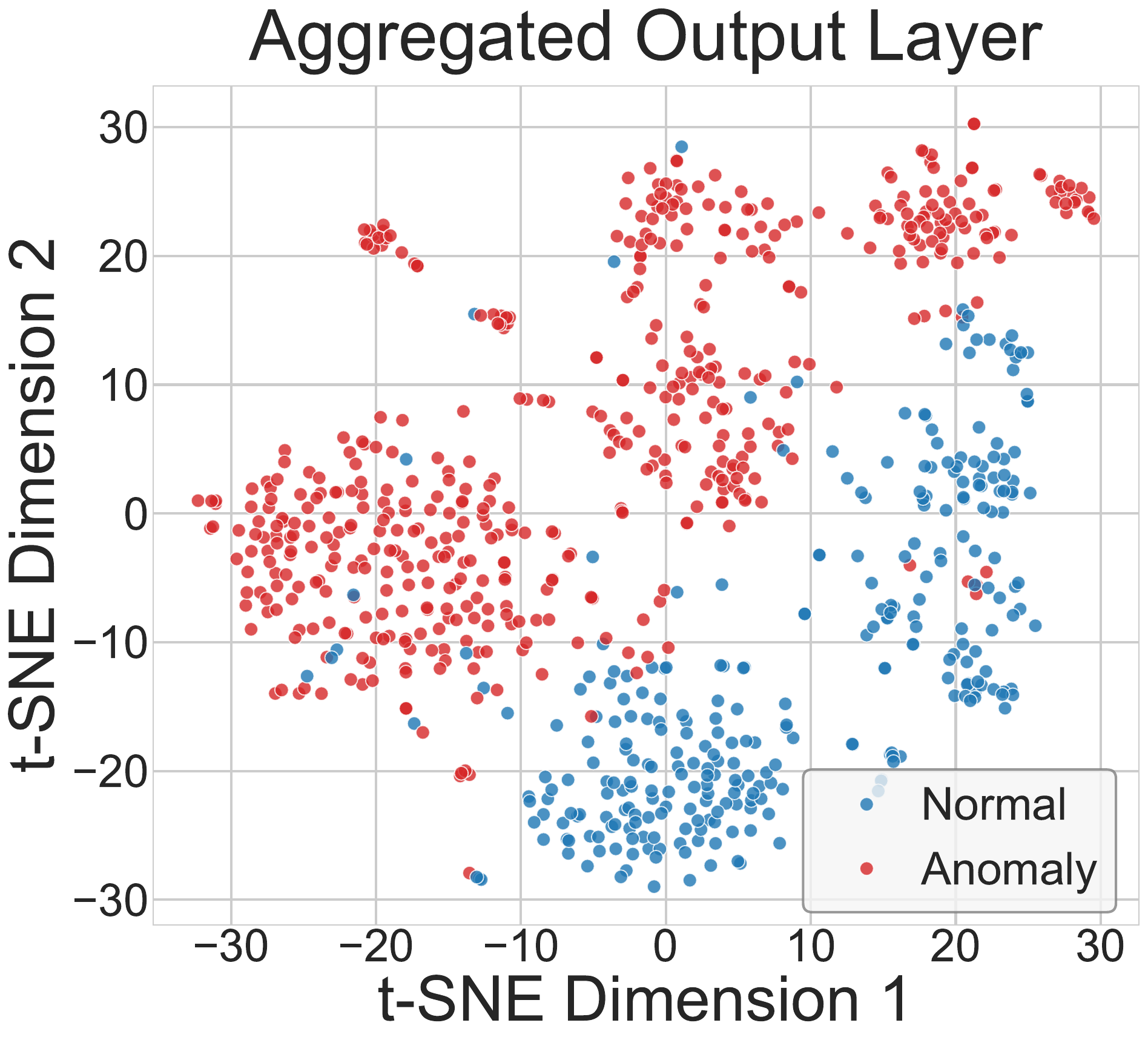}
        \caption{Features from Full Layer}
        \label{fig:tsne_all}
    \end{subfigure}
    \begin{subfigure}[b]{0.49\columnwidth}
        \includegraphics[width=\textwidth]{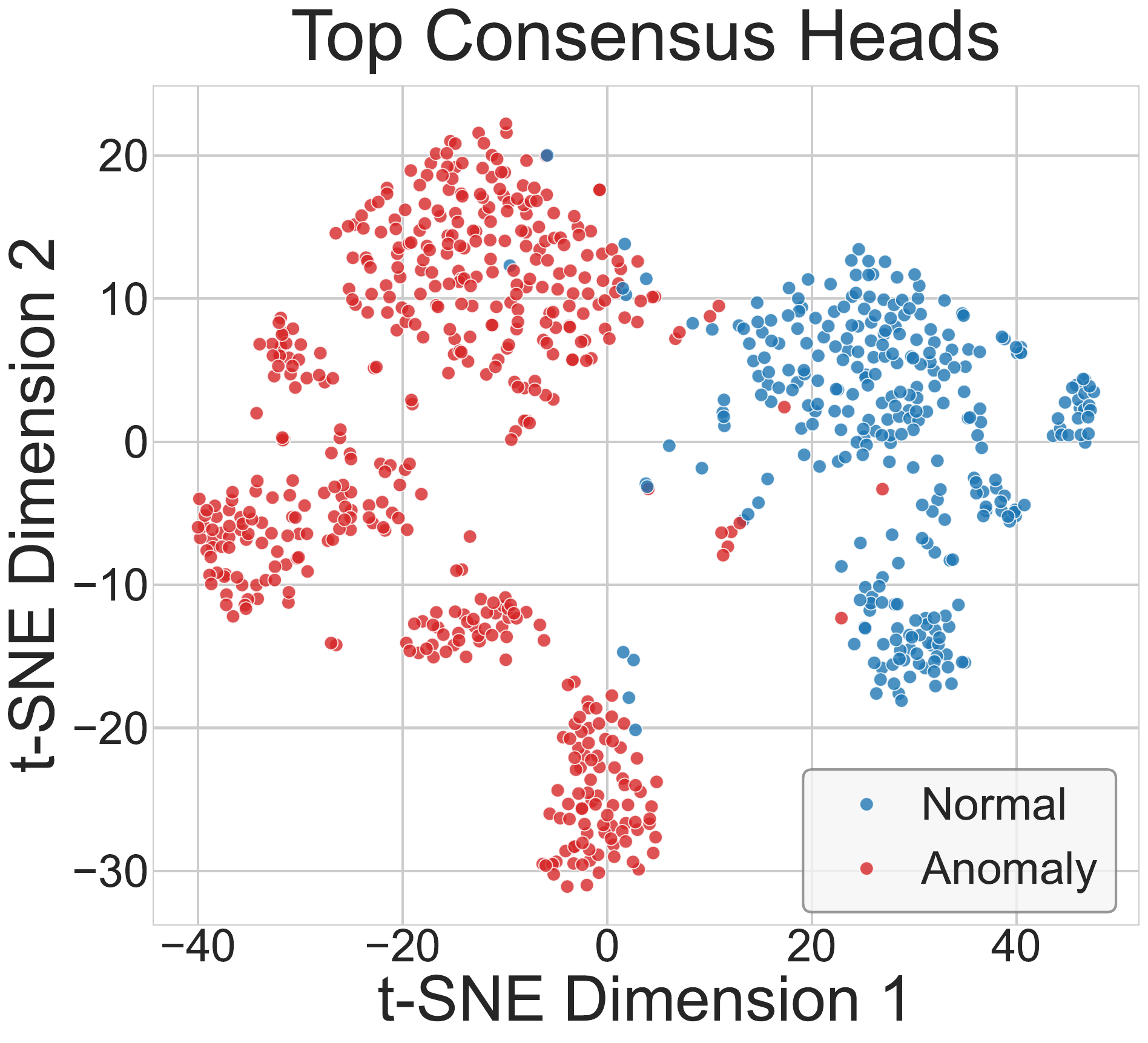}
        \caption{Features from Expert Heads}
        \label{fig:tsne_our}
    \end{subfigure}
    \caption{t-SNE visualization. (a) Full output layer show class overlap due to representation dilution. (b) Features from HeadHunt-VAD form distinct, separable clusters.}
    \label{fig:tsne_comparison}
\end{figure}

Figure~\ref{fig:tsne_comparison} visualizes the feature distributions of normal and abnormal samples using t-SNE. Features from the full output layer exhibit significant class overlap due to representation dilution. In contrast, features from our expert heads form clearly separated and compact clusters, confirming their superior discriminative power and better separability for anomaly detection.

\subsubsection{Qualitative Visualization}
\begin{figure}[t!]
    \centering
    \begin{subfigure}{1\columnwidth}
        \includegraphics[width=\linewidth]{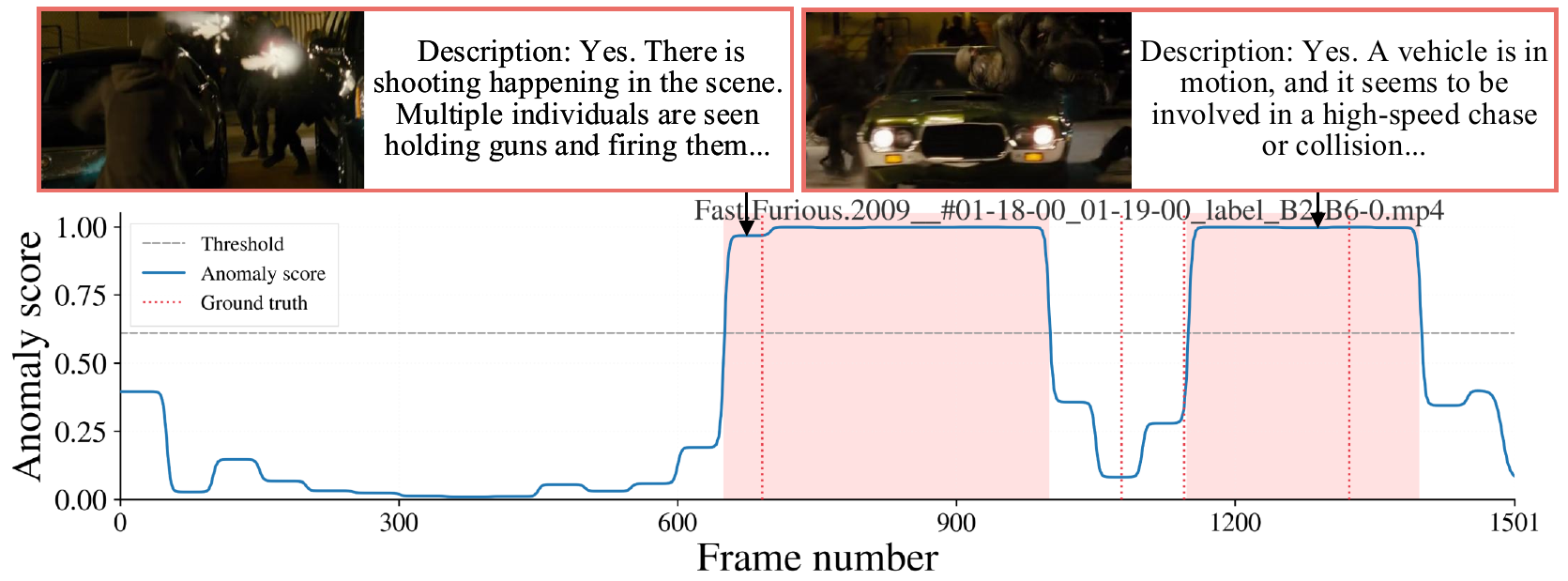}
        \label{fig:subfig1}
    \end{subfigure}
    \begin{subfigure}{1\columnwidth}
        \includegraphics[width=\linewidth]{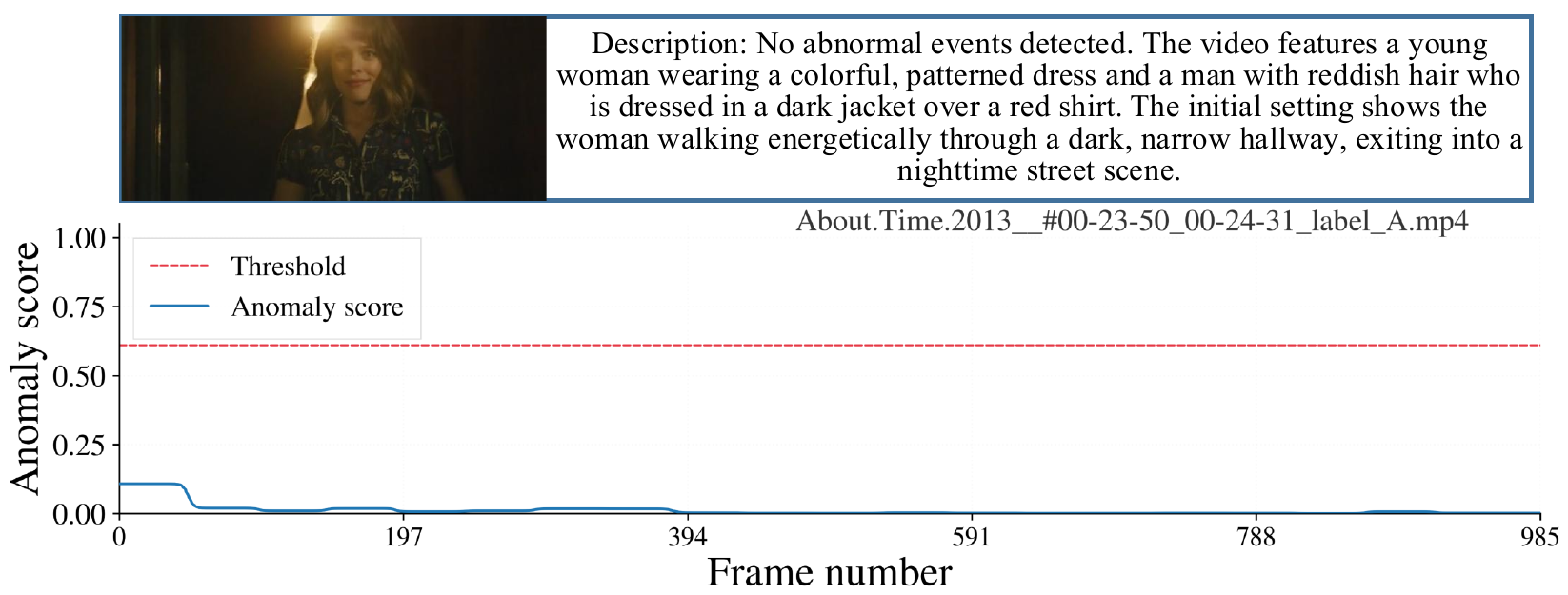}
        \label{fig:subfig3}
    \end{subfigure}
    \caption{Qualitative results of HeadHunt-VAD on XD-Violence. Each panel shows a representative video snippet and the anomaly curve. The shaded regions denote the detected anomaly frames that surpasses the threshold. Further generated descriptions are also provided.}
    \label{fig:qualitative_results}
\end{figure}

Figure~\ref{fig:qualitative_results} shows qualitative results from XD-Violence. For each video, the plot
shows the anomaly curves across different frames. For abnormal video, the frame-level probability curve generated by our method rises sharply and aligns precisely with the ground-truth temporal segments. Conversely, the anomaly curve of normal video remains consistently low and stable. These results demonstrate that HeadHunt-VAD provide reliable and accurate temporal localization of anomalous events. More visualization are provided in the supplementary materials.

\section{Conclusion}
\label{sec:conclusion}
In this paper, we introduced HeadHunt-VAD, a novel tuning-free paradigm that moves beyond the limitations of lossy textual outputs and diluted layer-level representations in MLLM-based VAD. Our method proactively identifies and leverages a sparse set of robust, anomaly-sensitive attention heads within a frozen MLLM, pinpointed by a multi-criteria Robust Head Identification module. Features from these expert heads are then channeled into lightweight modules for highly efficient scoring and localization. Extensive experiments on two major benchmarks demonstrate that HeadHunt-VAD establishes a new state-of-the-art among tuning-free methods, achieving superior performance with remarkable computational and data efficiency. Our work validates head-level probing as a powerful and practical paradigm for real-world video anomaly detection.


\bibliography{headhunt}

\end{document}